\documentclass[10pt,twocolumn,letterpaper]{article}

\usepackage[pagenumbers]{cvpr} 

\usepackage{graphicx}
\usepackage{amsmath}
\usepackage{amssymb}
\usepackage{booktabs}

\usepackage[pagebackref,breaklinks,colorlinks]{hyperref}

\usepackage[capitalize]{cleveref}
\crefname{section}{Sec.}{Secs.}
\Crefname{section}{Section}{Sections}
\Crefname{table}{Table}{Tables}
\crefname{table}{Tab.}{Tabs.}

\def\cvprPaperID{*****} 
\def\confName{CVPR}
\def\confYear{2022}

\newif\ifcomments

\usepackage{multirow}
\usepackage{cuted}
\usepackage{makecell}

\newcommand{\name}{X\&Fuse\xspace}
\newcommand{\method}{Fuse\xspace}

\begin{document}

\title{X\&Fuse: Fusing Visual Information in Text-to-Image Generation}

\author{Yuval Kirstain
\and
Omer Levy \\  \vspace{-0.1cm} \\
Tel-Aviv University \\
  \small{\texttt{yuval.kirstain@cs.tau.ac.il}}\vspace{-1.0cm}
\and
Adam Polyak
}

\maketitle


\begin{strip}\centering
  \centering
  \begin{tabular}{c@{~~}c@{~~}c}
   \textbf{Retrieve\&\method} & 
   \textbf{Crop\&\method} &
   \textbf{Scene\&\method} \\
   
  \includegraphics[height=0.45\paperwidth,trim={2 0 0 0},clip]{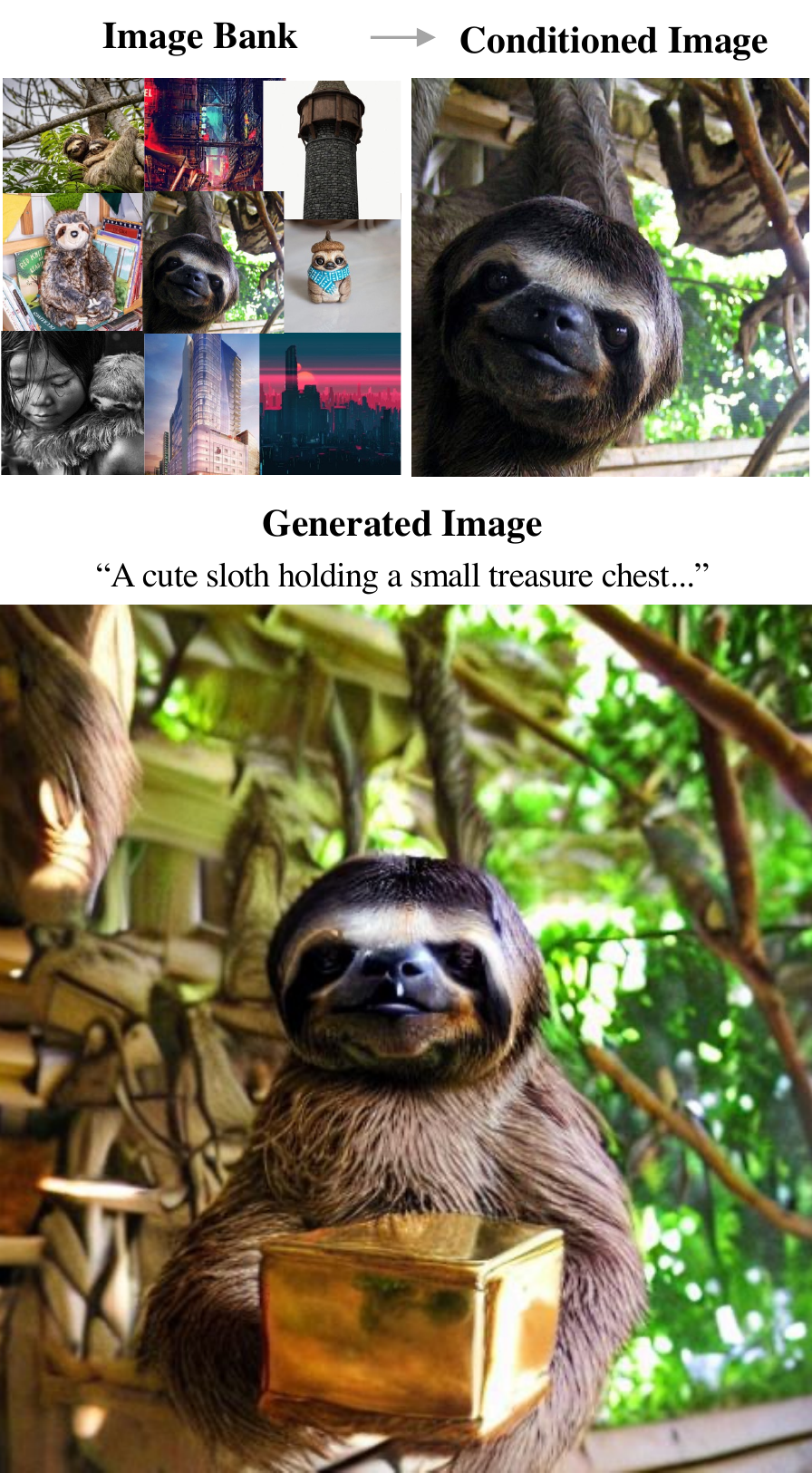} & 
  \includegraphics[height=0.45\paperwidth,trim={2 0 0 0},clip]{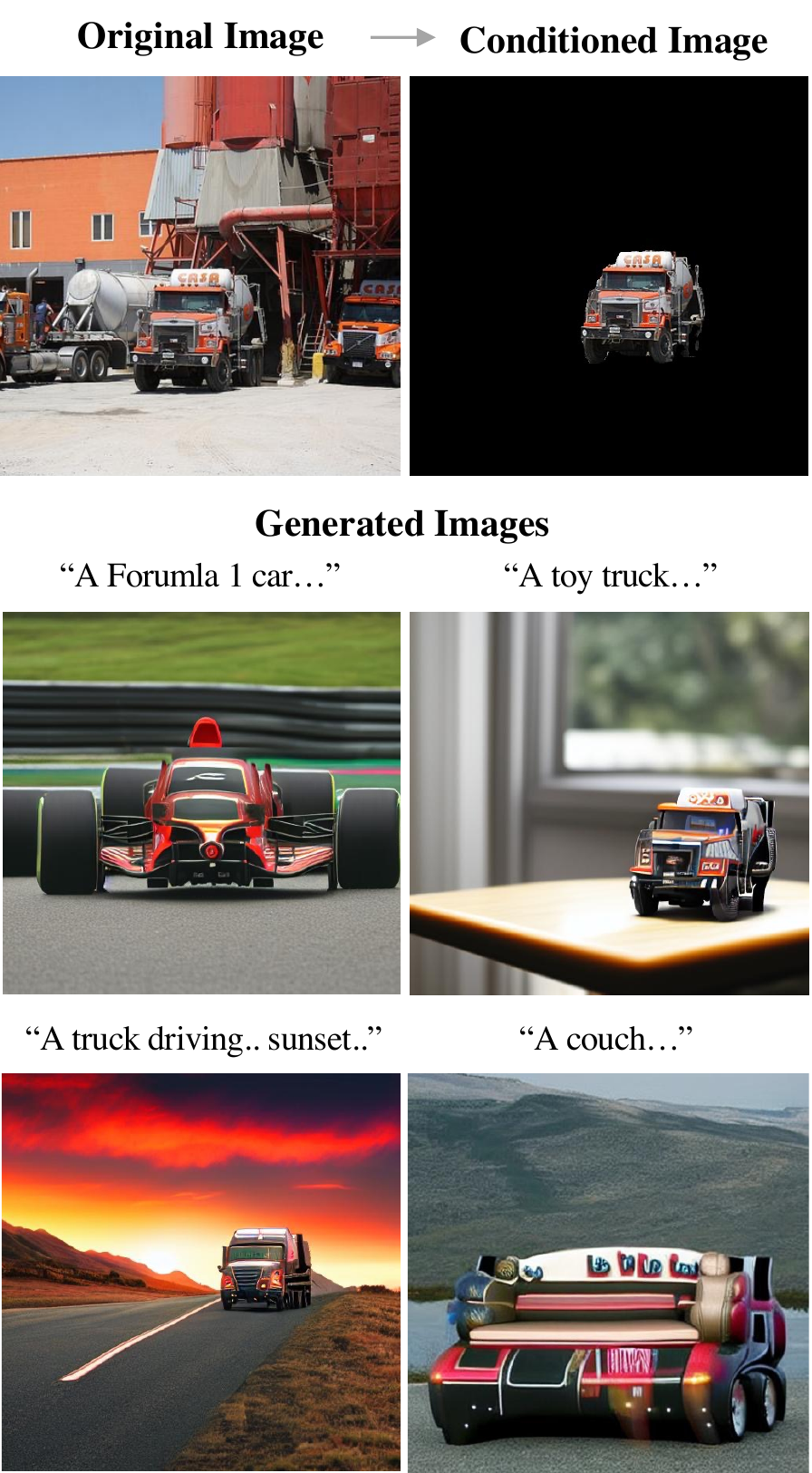} & 
  \includegraphics[height=0.45\paperwidth,trim={2 0 0 0},clip]{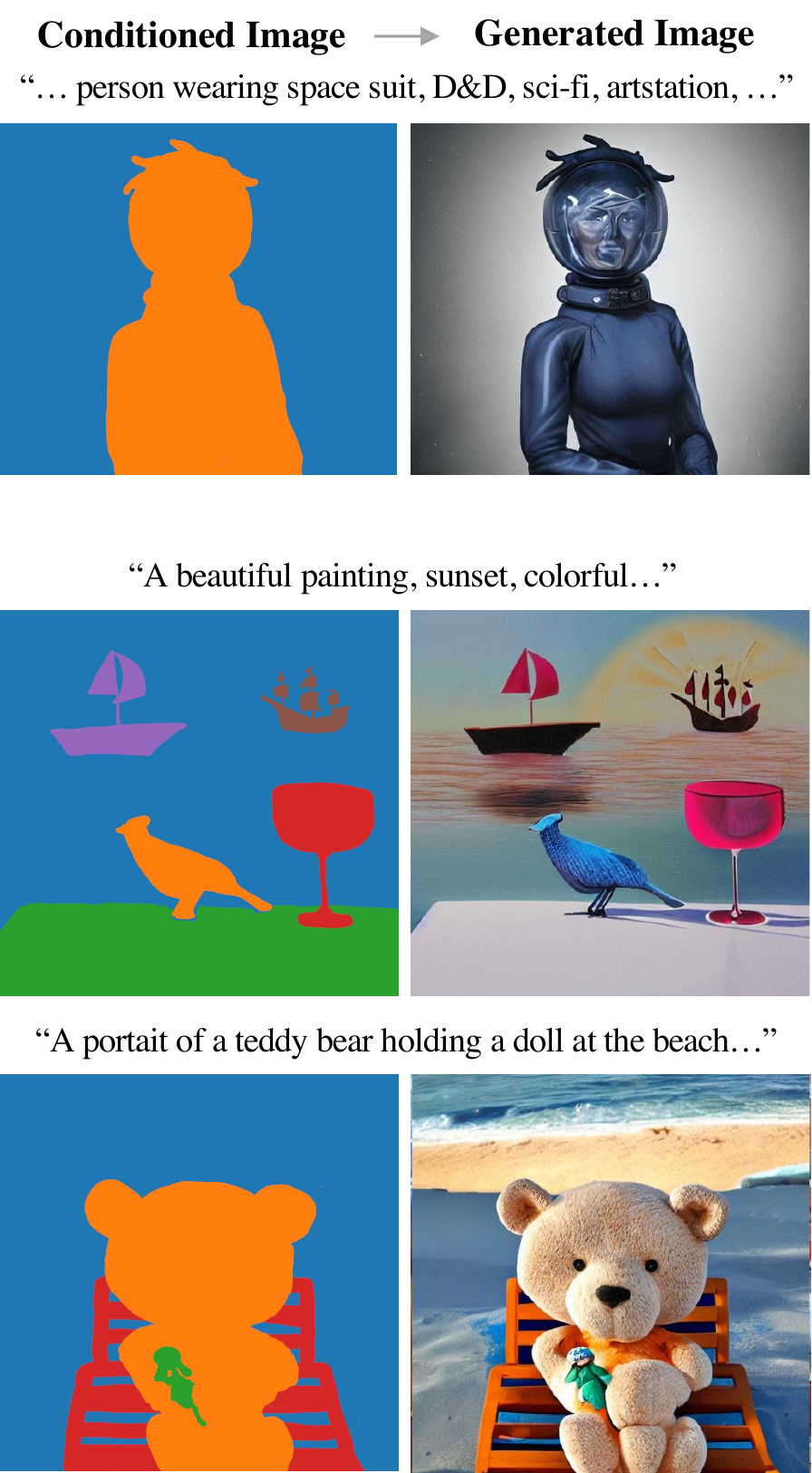} \\
  (a) & (b) & (c)
  \end{tabular}
  \vspace{+0.3cm}
  \captionof{figure}{\name conditions the model on both visual and textual information. (a) Top row: the bank of images (left) and the retrieved image (right). (b) Top row: the original image (left) and the (cropped) subject image (right). The bottom four images were generated using the subject image and corresponding caption. (c) Each row shows the caption, scene image (left) and the generated image (right).}
  \vspace{+0.3cm}
  \label{fig:teaser}
\end{strip}

\begin{abstract}
We introduce \name, a general approach for conditioning on visual information when generating images from text.
We demonstrate the potential of \name in three different text-to-image generation scenarios.
(i) When a bank of images is available, we retrieve and condition on a related image (Retrieve\&Fuse), resulting in significant improvements on the MS-COCO benchmark, gaining a state-of-the-art FID score of 6.65 in zero-shot settings.
(ii) When cropped-object images are at hand, we utilize them and perform subject-driven generation (Crop\&Fuse), outperforming the textual inversion method while being more than $\times100$ faster.
(iii) Having oracle access to the image scene (Scene\&Fuse), allows us to achieve an FID score of 5.03 on MS-COCO in zero-shot settings. 
Our experiments indicate that \name is an effective, easy to adapt, simple, and general approach for scenarios in which the model may benefit from additional visual information. 
\end{abstract}

\section{Introduction}
\label{sec:intro}

Current state-of-the-art text-to-image generation diffusion models are restricted to textual inputs~\cite{dalle2,imagen,parti}. Approaches that do use visual inputs~\cite{knn-us,knn-them} (1) do not attempt to challenge their text-based counterparts as they are usually focused on textless training, and (2) limit themselves to a CLIP~\cite{clip} embedding to represent the image. Therefore, they risk information loss and are unable to process the conditioned image, in a manner that is unique to the example at hand.

In this paper, we present a new general approach, \name, to utilize \textit{visual information} on top of the textual information. \name (1) can fit various scenarios, (2) is able to process \textit{all} of the visual information rather than just a small bottleneck representation, (3) enables full attention between different elements of the conditioned image, generated image, and text embeddings, and (4) can be easily applied to pretrained text-to-image models.
To demonstrate the potential of our approach, we explore three different text-to-image \textit{scenarios} (\cref{fig:teaser}).

First, we examine our ability to augment a text-to-image model with a retrieval component (Retrieve\&\method).
By finetuning the recently released Stable Diffusion model~\cite{ldm} with our approach, we gain state-of-the-art zero-shot FID~\cite{heusel2017gans} results on the MS-COCO benchmark~\cite{Lin2014MicrosoftCC} of $6.65$, surpassing larger and more compute demanding models such as Imagen~\cite{imagen} ($7.27$), DALL$\cdot$E-2~\cite{dalle2} ($10.39$), and Parti~\cite{parti} ($7.23$).
Additionally, we show that alternative methods for fusing visual information into the generation process (e.g. using CLIP~\cite{clip} image embeddings) do not yield meaningful improvements (\cref{sec:text-to-image}).

Second, we experiment with the recently proposed scenario of subject-driven generation~\cite{ti,dreambooth}, that requires the model to generate an image that is faithful to both visual properties of a subject, and a caption.
As there is no benchmark for properly evaluating subject-driven generation models, we create the first dataset for this task, SubGen (see \cref{subsec:subject-dataset,fig:subject-driven-comparison}). 
We then utilize cropped images to train our model on subject-driven generation in a self-supervised fashion. 
This yields a model that is preferred by human raters when compared with the textual inversion method~\cite{ti} (\cref{fig:subject-human-comparison}), while being more than $\times100$ faster, as it alleviates the textual inversion requirement of \textit{learning} the new subject at inference time (\cref{sec:subject-driven}). 

Third, is the scenario where the user can supply an image scene~\cite{make-a-scene}. 
We show that even though one may not expect for \name to be relevant for scene-based generation, by creating a plain RGB image from the scene information, and using it as the conditioned image (Scene\&\method), our model achieves an impressive FID score of 5.03 on MS-COCO+Scene~\cite{Lin2014MicrosoftCC,make-a-scene} (\cref{sec:scene}).

To conclude, \name is a general approach for conditioning on visual information, that is both effective in various scenarios, and can easily be applied to pretrained models.
As a result, \name creates new opportunities for injecting visual cues, when generating images from captions, with both immediate performance gains and potential applications for controlled image generation.

\section{Related Work}
\label{sec:related}
\vspace{-0.2cm}
\paragraph{Text-to-Image Generation}
State-of-the-art text-to-image generation models are improving in a rapid pace, mainly as a result of the (i) emergence of new large-scale training datasets~\cite{laion,dalle,imagen}, (ii) adaptation of large scale models~\cite{dalle,imagen,parti}, and (iii) rise of new modeling approaches~\cite{dalle,ldm,dalle2}.
However, those modeling approaches often restrict the model to textual inputs~\cite{dalle,ldm,dalle2,imagen,parti}. In this work, we focus on the potential of utilizing additional visual inputs, and suggest a general approach, suitable for~\textit{various} text-to-image scenarios that may benefit from such visual cues.

\smallskip\noindent{\bf Retrieval Augmented Diffusion Models\quad} 
Several works tried to augment the generation process of diffusion models with retrieved images~\cite{knn-us,knn-them}. However, unlike our work, they mostly use retrieval to facilitate textless training, or influence the style of an image~\cite{Rombach2022TextGuidedSO}, rather than improving results over text-based training.
Moreover, these methods use CLIP~\cite{clip} embeddings to represent the conditioned image. Thus, they are both limited by the bottleneck representation that may lose information, and are processed independently of the text, noised image, and even the task at hand.
We, on the other hand, suggest an approach that alleviates those shortcomings, and therefore, are able to improve performance over models that are restricted to textual inputs (see \cref{para:method-advanteges} and \cref{tab:alternative-ablation}).

\smallskip\noindent{\bf Subject-Driven Generation\quad} 
Recently, \cite{ti,dreambooth} introduced the task of subject-driven generation, and used textual inversion~\cite{ti,dreambooth} to tackle the task. Roughly speaking, in textual inversion, an embedding vector that represents a specific object is \textit{learned}. Then, the learned vector is used to generate new images with properties derived from the set of images.
While \cite{ti,dreambooth} presented impressive qualitative results, textual inversion requires learning the new subject at inference time, which may take more than fifteen minutes for a single subject.
On the other hand, we pretrain our model on the task at hand, and therefore require only a few seconds to generate a new subject image.

\smallskip\noindent{\bf Scene-Based Generation\quad} 
Scene-based generation was first introduced by \cite{make-a-scene}, as a text-to-image scenario that additionally enables the user to specify the scene (layout and objects) of the generated image, and therefore, offers much more fine-grained control over the generated image. 
Unlike previous work, we condition our model on an \textit{RGB image} of the scene, and showcase the applicability of \name as a general approach that can be used to condition on different types of visual information. 
\section{Method}
In this section, we first explain how \name works (\cref{subsec:modeling}), and then describe some of its advantages when compared to other modeling alternatives (\cref{subsec:advanteges}).  

\label{sec:method}
\begin{figure}[t]
    \centering
    \setlength{\tabcolsep}{2.0pt}
    
    \begin{tabular}{cc}
        \includegraphics[width=0.47\textwidth]{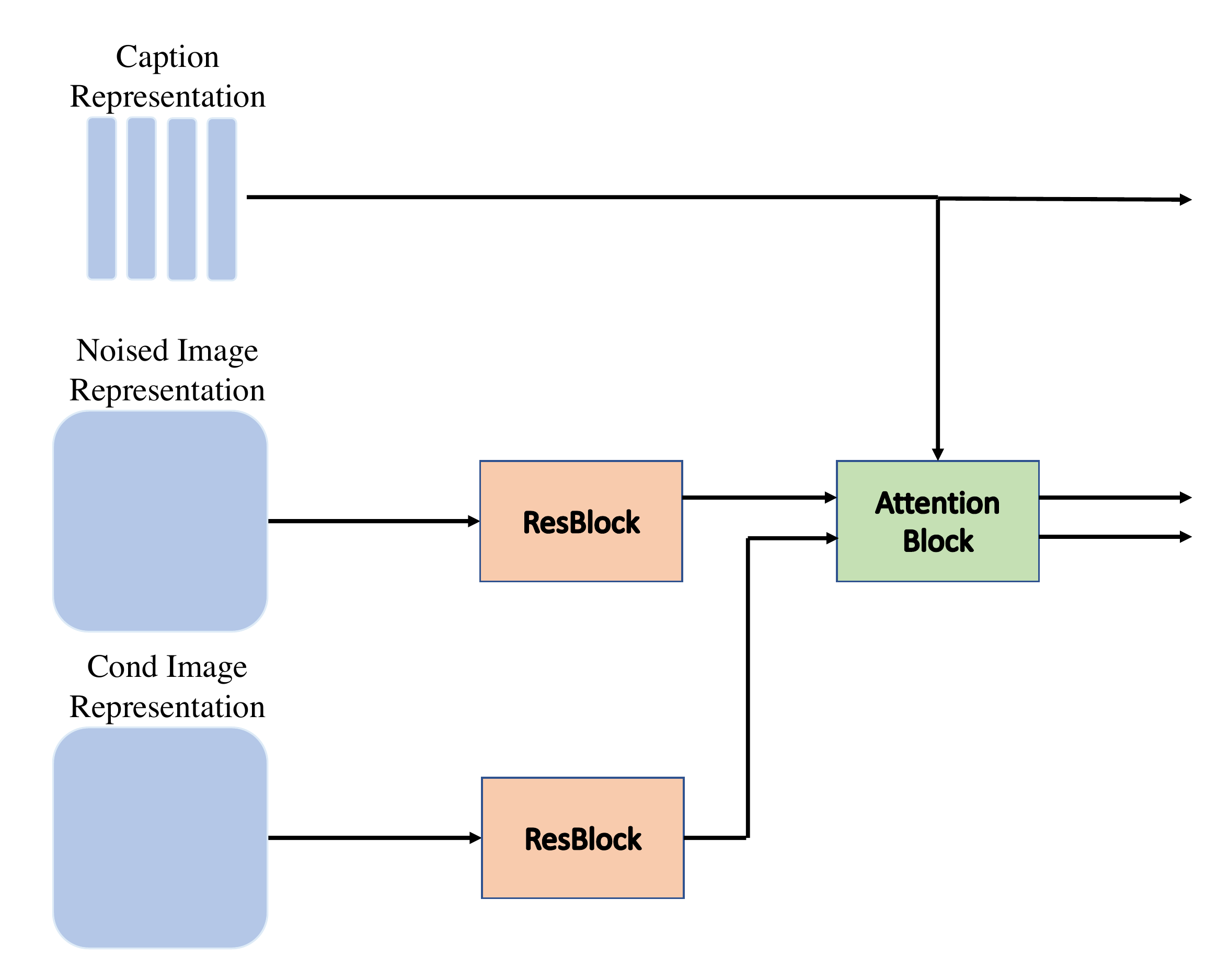} \\
    \end{tabular}
    
    \caption{\name high-level architecture. The conditioned image and noised image representations are processed separately using the \textit{same} ResBlocks and then concatenated before most attention blocks.}
    \label{fig:fusenet}
\end{figure}

\subsection{Modeling}\label{subsec:modeling}

In standard diffusion-based text-to-image generation, the U-Net architecture~\cite{Ronneberger2015UNetCN} receives two inputs: text embeddings, and a noised image. It then processes them using its two main building blocks: ResBlocks~\cite{resnet} and attention blocks~\cite{attention}.
The ResBlocks are operating on the noised image only, while the attention blocks enable interaction between different parts of the noised image via self-attention, and between the noised image and text embeddings via cross-attention.

Our method works similarly to the standard practice, with two main differences. Here, the model receives an \textit{additional} image (the conditioned image) to fuse into the generated image.
The conditioned image goes through the ResBlocks as if it was a noised image. 
However, prior to each attention block, we concatenate the conditioned image to the noised image, and allow interaction between the two via self-attention. 
Afterward, we separate the conditioned image representations from the noised image representations, to repeat the process in the following blocks.
Importantly, the weights that process the conditioned image and noised image are shared, which allows pretrained models to quickly adapt to the new visual inputs.
See \cref{fig:fusenet} for a visualization of the process.

\subsection{Advantages}\label{subsec:advanteges}
The \name method has several advantages over alternative methods, namely:

\smallskip\noindent{\bf Robustness to Spatial Differences\quad} 
\label{para:method-advanteges}
In \name, the conditioned image embeddings and noised image embeddings perform self-attention together. Therefore, \name does not assume spatial correspondence between the conditioned image and generated image. This is in contrast to concatenating the conditioned image along the channels dimension as commonly done in super-resolution models~\cite{Nichol2021ImprovedDD,Saharia2022ImageSV,dalle2,imagen} for example. Hence, it can fit a wider variety of scenarios and is more suitable for conditioning on retrieved images, object images, etc.

\smallskip\noindent{\bf No Information Loss\quad} 
\name provides the model with access to \textit{all} of the visual information that the conditioned image withholds. Therefore, our model is able to preserve identity in subject-driven generation. Do note, that a bottleneck representation of CLIP embedding, for example~\cite{knn-us, knn-them}, can potentially lose such fine-grained information.

\smallskip\noindent{\bf Example and Scenario Aware Processing\quad} 
\name learns how to process the conditioned image given the specific example and model inputs. This is in contrast to using a fixed representation extracted from a pretrained encoder such as CLIP. Therefore, our models are not limited by the modeling abilities of a pretrained model, and can be easily adjusted to new distributions, such as image scenes, or crops of objects; both are probably very different from the data used to train CLIP.

\smallskip\noindent{\bf Easy Adaptation to Pretrained Models\quad} 
Importantly, \name does not require any new weights. Thus, when initializing the model from a U-Net that was previously trained on text-to-image, it rapidly adapts to new tasks. This is unlike task specific architectures that require adding new weights and dropping existing ones. We consider this advantage to be especially important given the enormous amount of compute required to train state-of-the-art text-to-image generation models.


\begin{table}[t]
\centering
\small
\begin{tabular}{@{}lrrrrr@{}}
\toprule
\textbf{Model} & \textbf{FID}  $\downarrow$ & \textbf{CLIP-Score} $\uparrow$ \\
\midrule
DALL$\cdot$E2~\cite{dalle2} & 10.39 & -- \\
Imagen~\cite{imagen} & 7.27 & -- \\
Parti~\cite{parti} & 7.23 & -- \\
\midrule
SD~\cite{ldm} & 7.65 & \textbf{0.258} \\
SD (Cont) & 7.24 & 0.257 \\
\midrule
\textbf{Retrieve\&Fuse} & \textbf{6.65} & 0.253 \\
\bottomrule
\end{tabular}
\caption{Results on the MS-COCO benchmark. We present FID scores for all models, and CLIP-Score for SD~\cite{ldm} as it is publicly available, SD (Cont) and Retrieve\&\method.}

\label{tab:text-to-image}
\end{table}

\section{Vanilla Text-to-Image Generation}
\label{sec:text-to-image}

\begin{figure*}[!ht]
    \centering
    \setlength{\tabcolsep}{2.0pt}
    \begin{tabular}{ccc}
        \rotatebox{90}{\phantom{} DALL$\cdot$E-2 (prod)~\cite{dalle2}} &
        \includegraphics[width=0.98\textwidth]{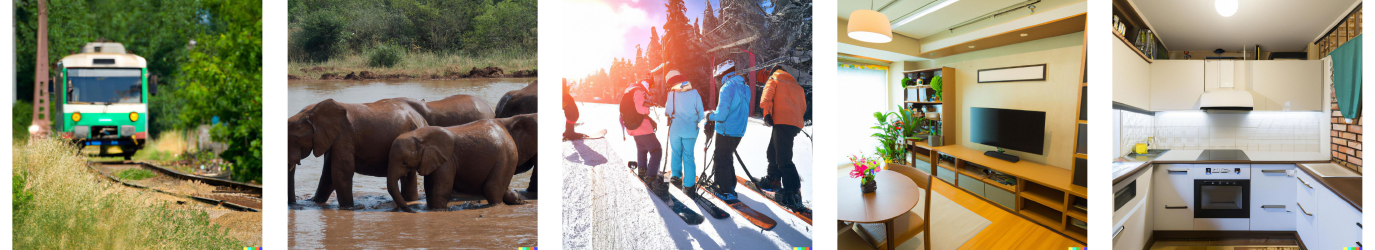} \\

        \rotatebox{90}{\phantom{AAAA} SD~\cite{ldm}} &
        \includegraphics[width=0.98\textwidth]{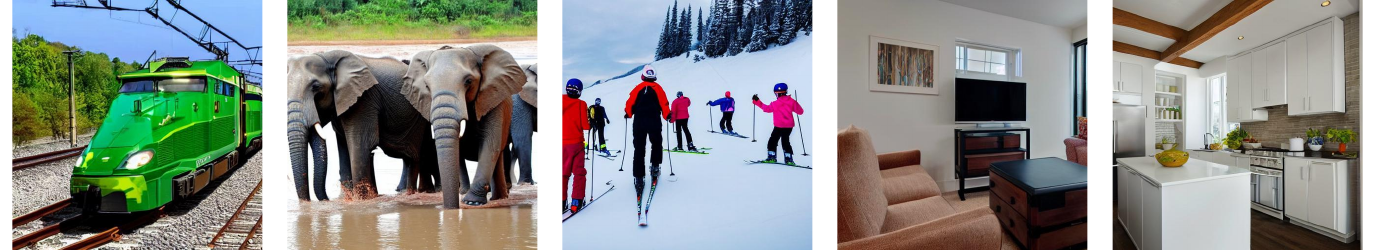} \\

        \rotatebox{90}{\phantom{A} \textbf{Ret\&Fuse}} &
        \includegraphics[width=0.98\textwidth]{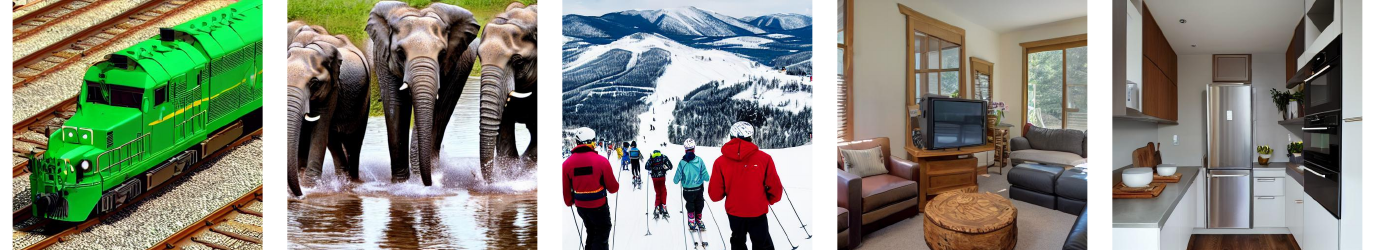} \\

    \end{tabular}
    \begin{tabular}{c@{\hskip 1.2cm}c@{\hskip 0.9cm}c@{\hskip 0.9cm}c@{\hskip 0.9cm}c@{\hskip 0.9cm}c@{\hskip 0.9cm}c}
        & \makecell{``a green train \\ is coming down \\ the tracks''}
        & \makecell{``a group of \\ elephants walking \\ in muddy water''}
        & \makecell{``a group of skiers \\ are preparing to ski \\ down a mountain''}
        & \makecell{``a living area \\ with a television \\ and a table''}
        & \makecell{``a small kitchen \\ with a \\ low ceiling''}
    \end{tabular}
    \caption{Qualitative comparison with the original Stable Diffusion model and DALL$\cdot$E-2 (prod.) on random images from MS-COCO.}
    \label{fig:text_to_image_qual}
\end{figure*}

In this section, we describe our experimental setup and experiments that involve the vanilla text-to-image scenario, in which the algorithm receives a caption as its only input. This includes the evaluation (\cref{subsec:text-to-image-eval}), Retrieve\&Fuse (\Cref{subsec:retrieve-and-fuse}), results (\cref{subsec:text=to-image-res}), and ablations (\cref{subsec:text-to-image-ablations}).

\subsection{Evaluation}\label{subsec:text-to-image-eval}

\smallskip\noindent{\bf Benchmark\quad} 
We follow the standard practice~\cite{make-a-scene,dalle2,imagen,parti}, and evaluate our results on $30k$ images from MS-COCO~\cite{Lin2014MicrosoftCC} validation set in zero-shot settings. That is, we evaluate the generalization ability of the different models to the MS-COCO validation set, without training them beforehand on the MS-COCO training set.

\smallskip\noindent{\bf Baselines\quad} 
\label{para:text-to-image-baselines}
We consider as baselines the released v1.3 variant of Stable Diffusion (SD)~\cite{ldm}, which is also the initialization point for our model.
Additionally, to compensate for the additional data, steps, and possible hyperparameter differences, we continue training the SD model with the exact same settings as our model, and report results for this model, which we refer to as SD (Cont).
Since Imagen~\cite{imagen}, DALL$\cdot$E-2~\cite{dalle2}, and Parti~\cite{parti} are not publicly released, we add comparable results that were reported in their original papers.

\smallskip\noindent{\bf Metrics\quad} 
We follow standard practice~\cite{make-a-scene,dalle2,imagen,parti} and measure FID-30K~\cite{heusel2017gans} and CLIP-Score~\cite{clip-score}.

\subsection{Retrieve\&Fuse}\label{subsec:retrieve-and-fuse}
The Retrieve\&Fuse approach is a special case of X\&Fuse, in which we \textit{retrieve} the conditioned image from a bank of images. Hence, this approach can be considered to be a semi-parametric one, that makes use of a large database of text-image pairs to augment the generated image.

\smallskip\noindent{\bf Conditioned Images\quad} 
We retrieve images by either using a text-index, or an image-index, both embedded by the ViT-L/14 CLIP variant~\cite{clip}.
Additionally, during training we encourage the model to use the conditioned images by conditioning $40\%$ of the time on images retrieved from the text input using the text index, another $40\%$ on images retrieved from the ground truth image CLIP embedding using the image index, and the remaining $20\%$ on the ground truth itself.
At inference time we only have the textual input at our disposal, yet can choose to retrieve from the text index, image index, or combine the two.
For simplicity, in our main experiments we only retrieve images with the text-index, and ablate this decision in~\cref{subsec:text-to-image-ablations}.

\smallskip\noindent{\bf Training Implementation Details\quad} 
\label{para:text-to-image-implementation}
We initialize our model from the v1.3 variant of Stable Diffusion (SD)~\cite{ldm}, use its noise scheduler, classifier-free guidance training ratio, and train it using Adam~\cite{kingma2017adam} with batches of $2,048$ examples each, a weight decay value of $0.01$, for \textit{only} fifty thousand steps. 
The learning rate is warmed up for one thousand steps to a maximum value of $10^{-4}$, after which it decays linearly to zero.
For the text-image dataset, we use the high-definition (HD) subset from~\cite{laion}. 
We also use two million examples from this dataset to create our index that we use during training.
We create the faiss~\cite{faiss} indices using the autofaiss repository\footnote{\url{https://github.com/criteo/autofaiss}}.
Last, we concatenate the conditioned image to the noised image before all attention layers in which the number of image embeddings is at most $1024$.

\smallskip\noindent{\bf Inference Implementation Details\quad} 
We do not perform hyperparameter tuning or optimize for FID, and simply use classifier-free guidance~\cite{classifierfree} with the best reported guidance scale for the Stable Diffusion model, which is $3$\footnote{See FID curve in \url{https://huggingface.co/CompVis/stable-diffusion-v1-3}.}. 
Additionally, we use DDIM~\cite{Song2021DenoisingDI} sampler with the standard value of $250$ steps~\cite{dalle2,imagen}. 
Thus, other guidance scale values or samplers may improve results. 
We take from the high-definition (HD) subset from~\cite{laion}, thirty-five million examples and use them for our bank of text-image pairs that we retrieve images from.

\subsection{Results}\label{subsec:text=to-image-res}
In \cref{tab:text-to-image}, we report the FID and CLIP-Score.
As can be seen, although our model (Retrieve\&\method) performs on-par in terms of CLIP-Score, it achieves superior results compared with the baselines in terms of FID, setting a new state-of-the-art result of 6.65; a better result than that of much larger and compute demanding baselines such as Parti~\cite{parti} (7.23) and Imagen~\cite{imagen} (7.27).
Interestingly, SD (Cont) improves as well compared with SD~\cite{ldm} by 0.41 FID. This may indicate that the slightly different hyperparameters we use (e.g. linearly decaying learning rate), are effective in improving performance.

\subsection{Ablation Study}\label{subsec:text-to-image-ablations}

\smallskip\noindent{\bf Alternative Conditioning Mechanisms\quad} 
\label{subsec:alternative}
Since we focus on the \name approach throughout the paper, it is unclear how alternative approaches for conditioning on visual data may perform.
Therefore, in this ablation study, we experiment with two different alternatives that we trained with the exact same settings as the Retrieve\&\method model. First, is the option of concatenating the retrieved image along the channels dimensions (Retrieve\&Channel), and concatenating a CLIP embedding of the image to the text embedding (Retrieve\&CLIP).
Additionally, we consider non-trainable alternatives: (1) the trivial null approach of returning the retrieved image (Retrieve\&Null), and (2) using the retrieved image as an ``init image''~\cite{Meng2022SDEditGI}, i.e. initializing the generation process after adding noise suitable for $t_0=0.05$ (Retrieve\&Init). 

As can be seen in \cref{tab:alternative-ablation}, while all \textit{trainable} methods gain some increased performance from the additional retrieved data, the \name method is the only one that gains clear improvements, and surpasses SD (Cont) by $0.6$ in terms of FID. 
As for the non-trainable methods, i.e. the Retrieve\&Null and Retrieve\&Init, they lead to degraded performance.
This also shows that the trainable methods learn more meaningful functions than merely copying.
All methods maintain similar CLIP-Score measures.

\begin{table}[t]
\centering
\small
\begin{tabular}{@{}lrr@{}}
\toprule
\textbf{Model} & \textbf{FID} $\downarrow$ & \textbf{CLIP-Score} $\uparrow$ \\
\midrule
SD (Cont) & 7.24 & \textbf{0.257}  \\
\midrule
Ret\&Null & 12.48 & 0.220 \\
Ret\&Init & 7.88 & 0.254  \\
\midrule
Ret\&Channel & 7.18 & 0.256 \\
Ret\&CLIP & 7.17 & 0.256 \\
\textbf{Ret\&Fuse} & \textbf{6.65} & 0.253  \\
\bottomrule
\end{tabular}
\caption{Comparison between image conditioning approaches on the MS-COCO benchmark.}
\label{tab:alternative-ablation}
\end{table}

\begin{figure}
    \centering
    \includegraphics[width=\linewidth]{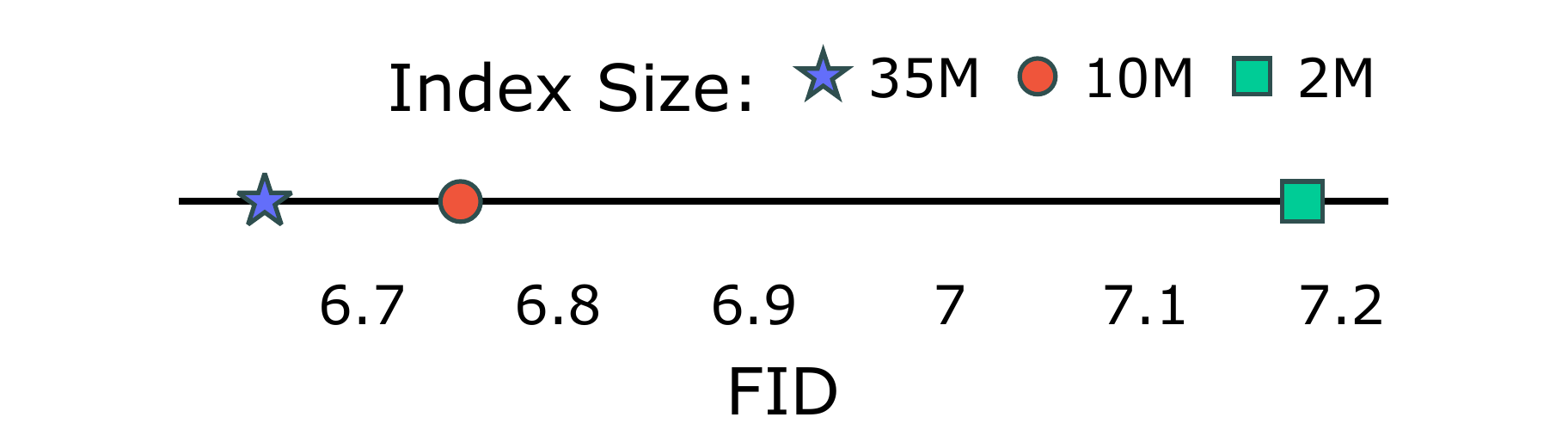}
    \caption{FID results on MS-COCO with indices of different size.}
    \label{fig:index-size-ablation}
\end{figure}

\smallskip\noindent{\bf Changing the Index Size\quad} 
As mentioned in \Cref{subsec:retrieve-and-fuse}, we train our model with an index of two million ($2M$) images, and during inference we use an index with $35M$ images. In this ablation, we investigate the effect of reducing the index size at inference time.
Specifically, we create an additional index with $10M$ entries, and calculate FID and CLIP-Score for each index size ($2M$, $10M$, $35M$).
While the CLIP-Score is similar regardless of the index size ($0.252-0.253$), as can be seen in \Cref{fig:index-size-ablation}, increasing the index size improves FID score. However, the performance margin when reducing the index size from $35M$ to $10M$ is much narrower (0.1) than when we reduce it to two million examples (0.53). 

\smallskip\noindent{\bf Removing the Ground Truth\quad} 
During training we train our model while $20\%$ of the time showing it the ground truth image, and in another 40\% of the time we use the ground truth image to retrieve an image which is similar to the ground truth image. 
In this ablation, we check whether these two heuristics are actually important to improve performance.
We do so by first training a model without showing it the exact ground truth image ($-$Ground Truth Exact), and then train another model without showing it the ground truth image, or an image that was retrieved by using the ground truth image ($-$Ground Truth Retrieval).
As can be seen in \cref{tab:show-ablation}, removing the ground truth conditioning yields slightly worse performance of $6.8$ FID, and the same applies when not using the ground truth at all to acquire the conditioned image ($6.92$ FID). 

\begin{table}[t]
\centering
\small
\begin{tabular}{@{}lrr@{}}
\toprule
\textbf{Model} & \textbf{FID} $\downarrow$ & \textbf{CLIP-Score} $\uparrow$ \\
\midrule
\textbf{Ret\&\method} & \textbf{6.65} & \textbf{0.253}  \\
$-$Ground Truth Exact & 6.80 & 0.252  \\
$-$Ground Truth Retrieval & 6.92 & 0.251  \\
\bottomrule
\end{tabular}
\caption{Ablation study: effect of not using ground truth conditioning.}
\label{tab:show-ablation}
\end{table}

\smallskip\noindent{\bf Changing the Index Type\quad} 
We check the effect of switching the index used at inference time to be based on image representations rather than text representations. 
We find that using an image-based index improves the CLIP-Score, from $0.253$ to $0.262$, but at the cost of a slight drop in FID, from $6.65$ to $6.82$.

\begin{figure}[t]
    \centering
    \setlength{\tabcolsep}{2.0pt}

    \begin{tabular}{cc}
        \includegraphics[width=0.48\textwidth]{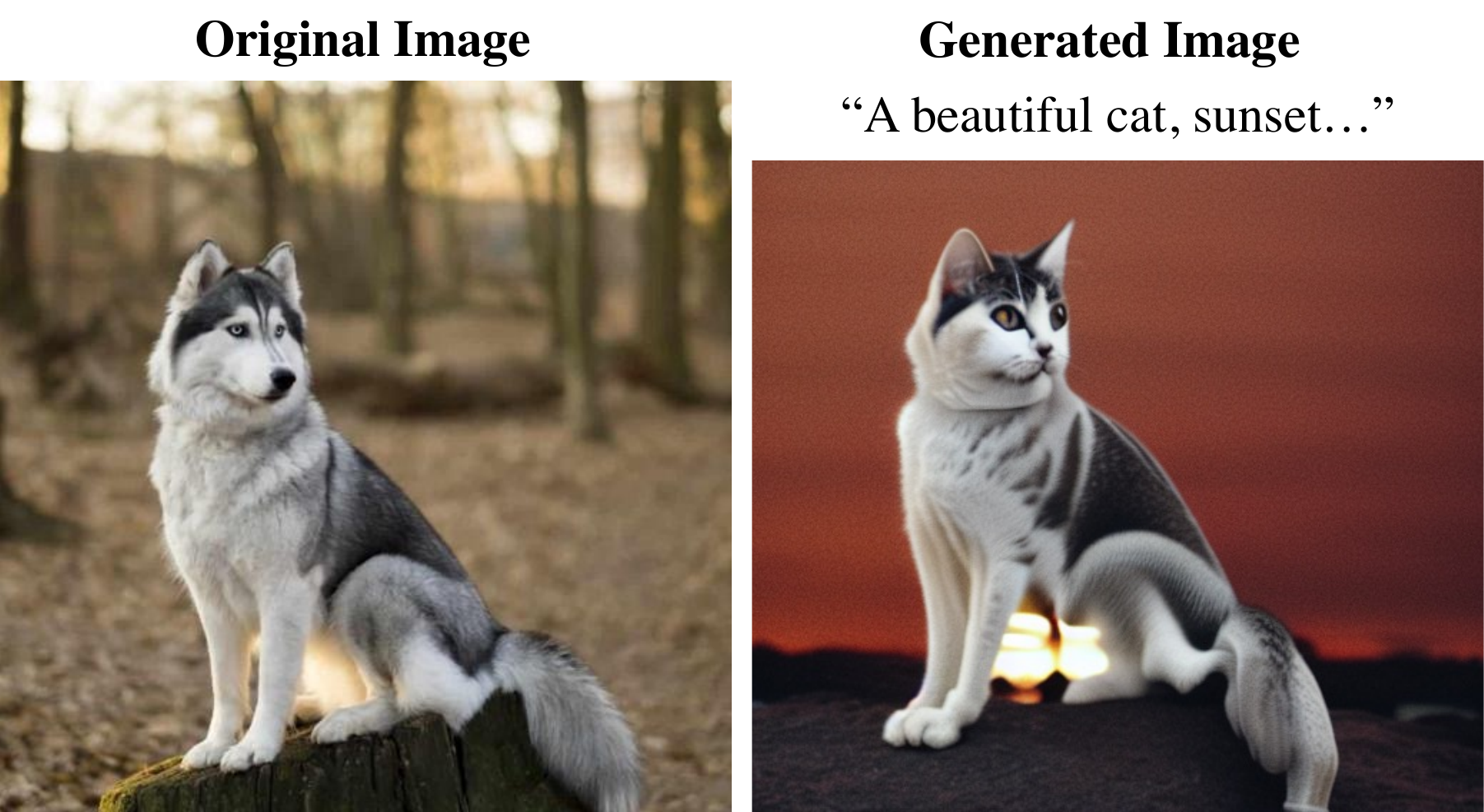} \\
    \end{tabular}
    
    \caption{An example generation from our Crop\&\method model. Additional samples are provided in \Cref{fig:crop-matrix}.}
    \label{fig:subject-driven-example}
\end{figure}

\section{Subject-Driven Generation}
\label{sec:subject-driven}

\begin{figure}[t]
    \begin{tabular}{cc}
        \includegraphics[width=0.44\textwidth]{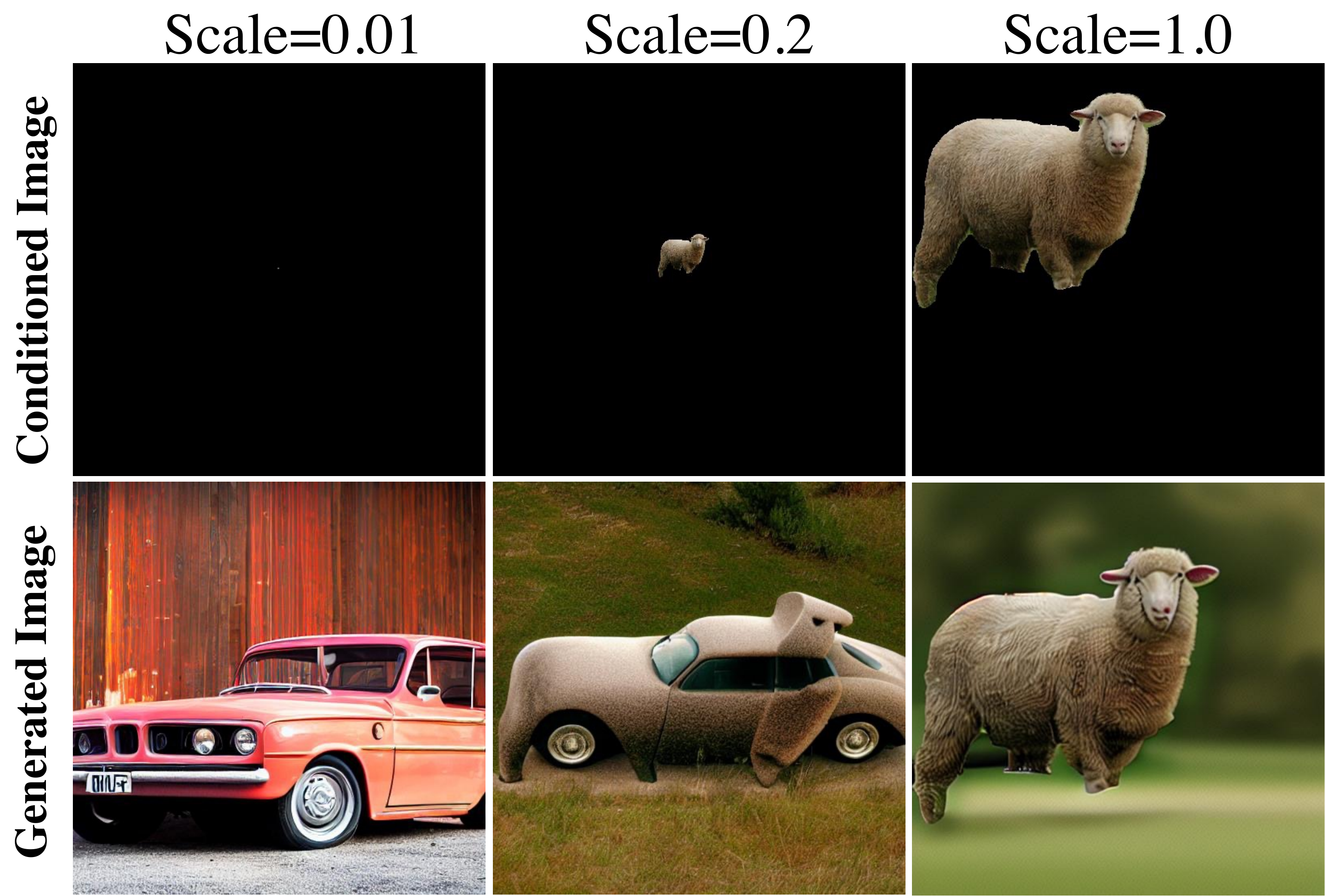} \\
    \end{tabular}
    \caption{Controlling text-alignment versus subject alignment with the scale parameter. All images are generated with the prompt ``A car.''.}
    \label{fig:subject-driven-scale}
\end{figure}

In subject-driven generation~\cite{ti,dreambooth} the model receives as input a textual description of an image, alongside \textit{visual} data to describe a \textit{subject}. 
The model is then expected to output an image that aligns with both the textual description and subject. 
Since there is no dataset to support quantitative results, in this section we first describe the dataset that we create for the task, which we call SubGen (\cref{subsec:subject-dataset}). 
We then follow with a description of our Crop\&\method approach for tackling the scenario (\Cref{subsec:crop-and-fuse}), report our results on the benchmark (\cref{subsec:subject-results}), and explain how we control the trade-off between faithfulness to the subject and text (\cref{subsec:subject-control}).

\subsection{Evaluation}\label{subsec:subject-dataset}

\smallskip\noindent{\bf Dataset Requirements\quad} 
We would like to create a dataset that emulates the scenario in which a user (1) has an image of a subject, (2) has an image caption that is related to the subject, and (3) expects the model to generate an image that aligns with both the subject and the caption.
Do note that while the desired output image should contain the subject, or key-properties of the subject, it might \textit{exclude} properties that are specific to the original image of the subject, such as the pose, lightning, position, etc.
Thus, when considering an example from MS-COCO, one cannot simply extract a crop of the subject from an original image and use the original caption.

\smallskip\noindent{\bf The SubGen Dataset\quad} 
To create our dataset, we initially take $5k$ images from the MS-COCO validation set. 
For each image, we use the segmentation model from~\cite{Carion2020EndtoEndOD} to \textit{identify} objects in the image.
Then, to collect high-quality subjects, we extract the object that was identified with the highest confidence. 

To create captions for the collected subjects, we leverage the in-context learning~\cite{gpt3} abilities of large language models~\cite{gpt3,jurrasic1,opt}.
Specifically, we take three random examples and create a demonstration\footnote{A demonstration is an example for the task that will be appended to the prompt used for in-context learning~\cite{lm-bff}.} from each one with the following template:
\texttt{Detected:~[Subject Type]~|~Caption:~[Caption]}.
By concatenating those three demonstrations, and prompting the model with the relevant subject type, we derive our final prompt.
This allows us to create a prompt for caption generation that \textit{reveals} the type of the object (e.g. dog), but \textit{conceals} details that are specific to the image it was taken from (like position or pose). 

Then, for each label, we sample five times from the Jurassic-J1~\cite{jurrasic1} language model a matching caption.
We then filter images with people, and subjects that are too small or too large, in order to keep subjects that can properly identified, which leaves us with about ten thousand images. 
We call this dataset SubGen.
Do note, that while~\cite{ti,dreambooth} show results that are based on multiple images for each object, SubGen only has one image per subject, which might make it more challenging for some approaches. 
We provide additional details and sampled examples in \Cref{fig:crop-matrix}.

\smallskip\noindent{\bf Metrics\quad} 
We consider three automatic metrics: (1) FID between generated images and real images from MS-COCO\footnote{We use images that that were not used to collect prompts or crops to prevent copying bias.}, (2) CLIP-Score between the caption and generated image, and (3) CLIP-Score between the subject crop and generated image. 
Additionally, we perform human evaluation and assess human preferences in terms of image quality, text-faithfulness, and faithfulness to the cropped subject image.

\smallskip\noindent{\bf Baselines\quad} 
The baseline methods selected are the vanilla Stable Diffusion (SD) model, SD (Cont) (see \cref{para:text-to-image-baselines}), and the textual inversion method~\cite{ti} applied with SD (TI).
To train TI, we use the implementation of Hugginface Diffusers repository\footnote{\url{https://github.com/huggingface/diffusers/tree/main/examples/textual_inversion}} and use their default hyperparameters. 
We use the type of the subject (e.g. dog) as the initialization token, and in cases where the type spans more than one token we simply use the first one.
However, because the textual inversion method is impractical to run at scale, as it requires more than fifteen minutes to learn a single subject, we only evaluate it using human evaluation on two hundred randomly sampled examples.

\subsection{Crop\&Fuse}\label{subsec:crop-and-fuse}

\smallskip\noindent{\bf Conditioned Image\quad} 
To train our model on subject-driven generation in a self-supervised fashion, we introduce a self-supervised scheme.
In this scheme we apply a segmentation model from~\cite{Carion2020EndtoEndOD} on the ground truth image, extract a crop from it, augment the crop, and then use the augmented crop as a conditioned image.
Intuitively, augmentations encourage the model to use the input caption when reconstructing the augmented object.
Importantly, unlike existing methods, our approach does not \textit{learn} new concepts at inference time, and therefore, is able to output an image for a new subject within seconds.
However, a limitation of our approach is that we do not separate between properties that are unique to the image (e.g. subject pose) and the identity of the subject.

\smallskip\noindent{\bf Implementation\quad} 
We use the same hyperparameters and settings described in \cref{para:text-to-image-implementation}.
As for the augmentations that we apply to the crop, we use a random affine transformation, with a scaling factor between $0.2$ and $4$, a translation factor of $0.3$, and a degree range of $0-180$. 
At inference time, we set the scale factor to $0.3$, the translation factor to $0$, and the degree range to $0$.
See \cref{subsec:subject-control} for an explanation of the effect of scaling the subject.

\subsection{Results}\label{subsec:subject-results}
In \cref{fig:subject-driven-example,fig:crop-matrix} we show examples generated with our model, and in \cref{tab:subject-automatic}, we display the quantitative results on the SubGen dataset.
As can be expected, when compared with models that are unaware of the subject, our model is able to get a higher (by $~0.03$) CLIP-Score between the generated image and the subject image, and slightly lower (by $~0.008$) CLIP-Score between the generated image and the text.
However, the Crop\&Fuse FID score is lower than the baseline. This is especially interesting when considering that in human evaluations (\cref{fig:subject-human-comparison}) human raters slightly preferred the quality and text faithfulness of SD (Cont), which is the expected result, as this model is not constrained by the subject.
When comparing our model with TI, we find that human raters prefer \name by a large margin in all parameters.
As noted before, it is possible that when given more than one subject image, TI is able to perform better.

\begin{table}[t]
\centering
\small
\begin{tabular}{@{}lrrrrr@{}}
\toprule
\textbf{Model} & \textbf{FID} $\downarrow$ & \textbf{CLIP (Text)} $\uparrow$ & \textbf{CLIP (Sub)} $\uparrow$ \\
\midrule
SD~\cite{ldm} & 25.4 & \textbf{0.253} & 0.672 \\
\midrule 
SD (Cont) & 24.5 & 0.252 & 0.673 \\
\midrule
\textbf{Crop\&Fuse} & \textbf{22.4} & 0.244 & \textbf{0.701} \\
\bottomrule
\end{tabular}
\caption{Automatic evaluation results on the SubGen dataset. ``CLIP'' refers to CLIP-Score, Text means text-alignment, and Sub is used for subject-alignment.}
\label{tab:subject-automatic}
\end{table}

\begin{table}[t]
\centering
\small
\begin{tabular}{@{}lrrrrr@{}}
\toprule
\textbf{Crop\&Fuse VS} & \textbf{Quality} & \textbf{Text} & \textbf{Subject} \\
\midrule
SD (Cont) & 55\% & 55\% & 30\% \\
TI & 40\% & 29\% & 26\% \\
\bottomrule
\end{tabular}
\caption{Human evaluation results on SubGen. In each line, we report the model name and percentage of human annotators that favored the relevant model against Crop\&Fuse. We measure preference in terms of quality, text-alignment (Text), and subject-alignment (Subject).}
\label{fig:subject-human-comparison}
\end{table}

\subsection{Analysis}

\paragraph{Controlling the Subject Faithfulness}\label{subsec:subject-control}
During training, we apply augmentations on the subject to encourage the model to change the subject accordingly to the prompt, rather than simply ``outpainting'' the image.
Intuitively, the more we augment the subject, the harder it should be to reconstruct it. Therefore, the model should find the input caption to be more useful when reconstructing the subject.
Consequently, the generated image will be more faithful to the input caption, perhaps even at the cost of faithfulness to the subject.
At inference time, we can use this insight, and in \cref{fig:subject-driven-scale} we show an example in which we take a caption, ``A car.'' that is very unrelated to the subject, a sheep, and demonstrate how changing the scale value indeed affects the generation process to preserve fidelity to either the caption or the subject.
While in our experiments we simply set the scale to a value of $0.3$, it may be that adapting the scale for each subject independently, or applying more sophisticated augmentations at training or inference time, will result in improved performance.

\begin{table}[t]
\centering
\small
\begin{tabular}{@{}lrrrrr@{}}
\toprule
\textbf{Model} & \textbf{FP} $\downarrow$ & \textbf{Param}
$\downarrow$ & \textbf{TF}
$\downarrow$  & \textbf{FID} $\downarrow$ & \textbf{CLIP} $\uparrow$ \\
\midrule
Make-A-Scene~\cite{make-a-scene} & 1024 & 4B & 43.9 & \textbf{4.69} & \textbf{0.262} \\ 
\midrule
\textbf{Scene\&Fuse} & \textbf{16} & 870M & 5.1 & 6.46 & 0.260 \\
\textbf{Scene\&Fuse} & 32 & 870M & 10.2 &  5.22 & 0.259 \\
\textbf{Scene\&Fuse} & 64 & 870M & 20.4 & 5.03 & 0.258 \\
\bottomrule
\end{tabular}
\caption{Results on the MS-COCO benchmark with scenes. We present number of forward passes required by each model to generate an image (FP), number of parameters (\#P), Tera floating-point operations (FLOPS) per generated image (TF), FID, and CLIP-Score (CLIP).}
\label{tab:scene-gen}
\end{table}

\begin{figure}[t]
    \centering
    \setlength{\tabcolsep}{2.0pt}

    \begin{tabular}{cc}
        \includegraphics[width=0.48\textwidth]{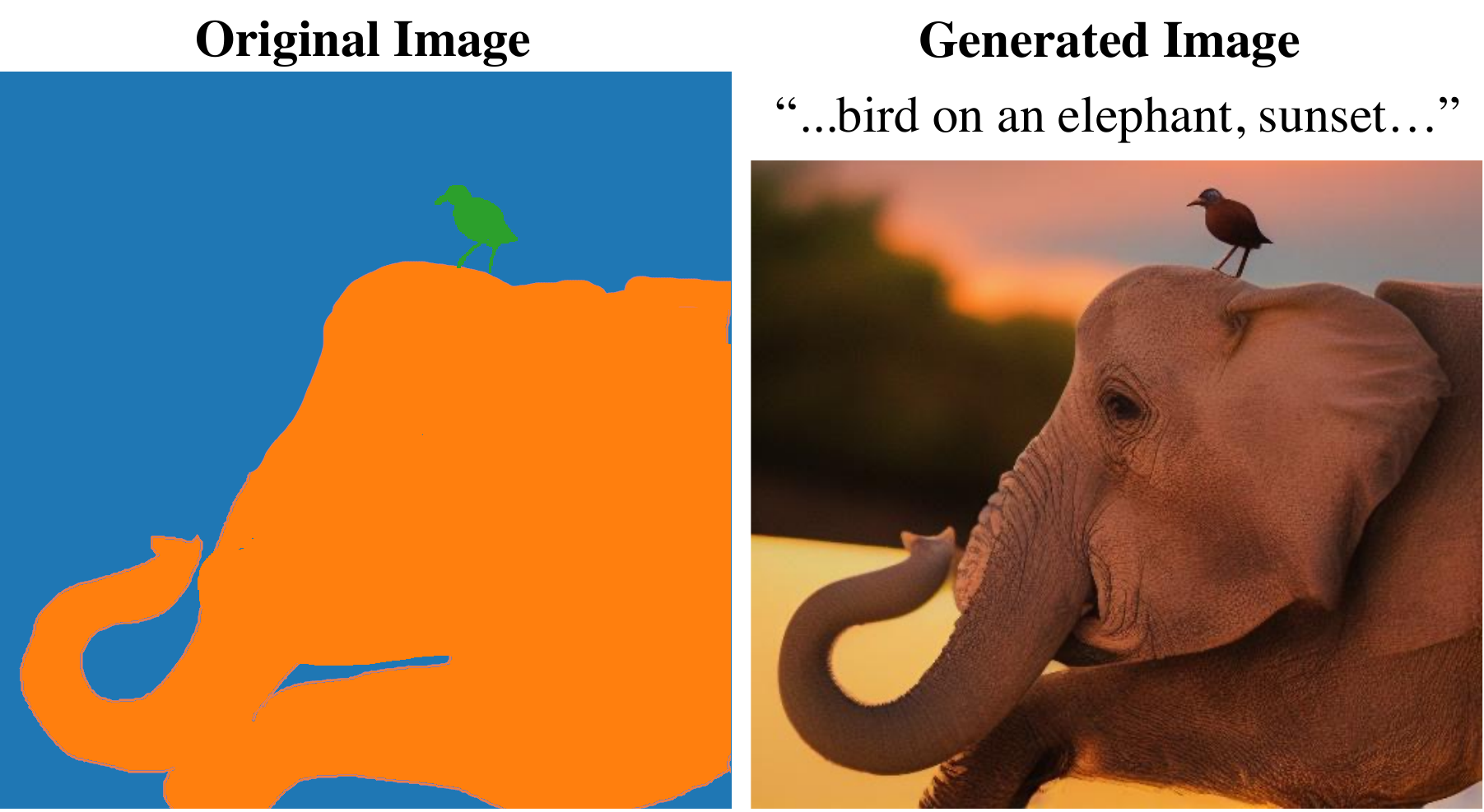} \\
    \end{tabular}
    
    \caption{An example generation from our Scene\&\method model. Additional samples are provided in \Cref{fig:scene-matrix}. }
    \label{fig:scene-driven-example}
\end{figure}

\section{Scene-Based Generation}
\label{sec:scene}
In scene-based generation~\cite{make-a-scene} the model receives an additional input that encodes the ``scene'' (information about the layout and objects in the image) of the image that it should generate. 
While intuitively there might be dedicated architectures that better fit this scenario, our goal here is not to achieve state-of-the-art results, but rather, to show the versatility and simplicity of the \name approach. 

\smallskip\noindent{\bf Conditioned Images\quad}
We condition our model on an RGB image of the scene. 
During training, we create the scene image by first applying a segmentation model on the ground truth image. Then, for each detected object, we add the predicted textual label to the caption and color the image according to its predicted mask.
During inference, we can \textit{color} the scene image, and add the objects description to the caption. 
We choose the color according to the order in which we add the textual description to the caption. 
For example, if we add "elephant, bird" to the prompt, the scene image should contain an orange elephant and green bird (see \cref{fig:scene-driven-example}). If we would switch the order of objects to "bird, elephant" we would also switch the colors.
This approach enables us to generate any object at inference time, because it does not assume a predefined set of object types. 
This is in contrast to the task-specific architecture of Make-A-Scene~\cite{make-a-scene} that does assume a predefined set of object types, and feeds their model with a tensor that contains for each pixel in the image a one-hot vector that represents the object type.

\smallskip\noindent{\bf Implementation\quad}
We use the same hyperparameters and settings described in \cref{para:text-to-image-implementation}, but use a higher guidance scale of $4$, a text dropout ratio of $10\%$ during training, and during classifier free guidance we always condition on the scene.
For a segmentation model, we use the default parameters from~\cite{Carion2020EndtoEndOD}.

\smallskip\noindent{\bf Benchmark\quad}
Similarly to \cref{sec:text-to-image}, we evaluate our results on the MS-COCO benchmark, but since we consider scene-based generation, we allow models to use a segmentation map of the ground truth image at inference time.

\smallskip\noindent{\bf Metrics\quad}
Similarly to \cref{subsec:text-to-image-eval}, we consider 30K-FID, and CLIP-Score as automatic metrics.

\smallskip\noindent{\bf Baselines\quad}
The baseline that we compare to at scene-based generation is Make-A-Scene. Do note that Make-A-Scene (1) is much more compute intensive than our model, as it has more than $\times4$ parameters than Stable Diffusion, (2) was trained for $170k$ steps, and its segmentation-specific VQGAN was trained for $600k$ iterations, while we finetune our model for only fifty thousand steps.

\smallskip\noindent{\bf Results\quad}
As we show in \cref{tab:scene-gen}, adding a scene input to our model enables us to obtain an FID score of 5.03 while using only 64 diffusion steps.
Interestingly, we notice that we are able to use a small number of diffusion steps without suffering from a large decrease in performance. Specifically, we are able to use only $32$ diffusion steps and attain FID score of 5.22, and $16$ diffusion steps yields FID score of 6.46. 
This is in contrast to Make-A-Scene, that indeed achieves an unprecedented FID score of 4.69, but is a $4B$ parameter models and requires $1024$ forward passes to output a single image, which makes our model substantially more efficient. For completeness, we report the number of floating-point operations per generated image  in \cref{tab:scene-gen}. We calculate the FLOPS for our model using the fvcore repository\footnote{\url{https://github.com/facebookresearch/fvcore}}, and for the transformer from Make-A-Scene, we use standard estimates derived from the models architecture and sequence length. 

\section{Conclusions}
\label{sec:conclusions}
In this paper, we introduced a new general approach, \name, for conditioning on visual information in the task of text-to-image.
We showcased the potential of the approach against strong baselines and modeling alternatives, and experimented with three scenarios that are very different from one another. 
\name has set a new state-of-the-art FID result in text-to-image, and showed impressive performance regardless of the scenario.
Thus, offering an appealing option for other scenarios that may benefit from additional visual information when generating images.


{\small
\bibliographystyle{ieee_fullname}
\bibliography{egbib}
}

\appendix

\section{SubGen}\label{ap:subgen-prompt}
We used the following prompt to create the captions in the SubGen dataset:
\texttt{"""Detected: bowl | Caption: A table with pies being made and a person standing near a wall with pots and pans hanging on the wall.
==
Detected: toilet | Caption: A cluttered room with a sink, a toilet and in industrial mop bucket.
==
Detected: skateboard | Caption: A skateboarder flipping his board on a street.
==
Detected: [label] | Caption:"""}

Additionally, we filter objects that consist of less than $200$ pixels, or more than $900$ pixels, as we saw that this heuristic many times filters objects that are too small or too big to be properly identified.

\begin{figure*}[t]
    \centering
    \setlength{\tabcolsep}{2.0pt}
    
    \begin{tabular}{cc}
        \includegraphics[width=0.98\textwidth]{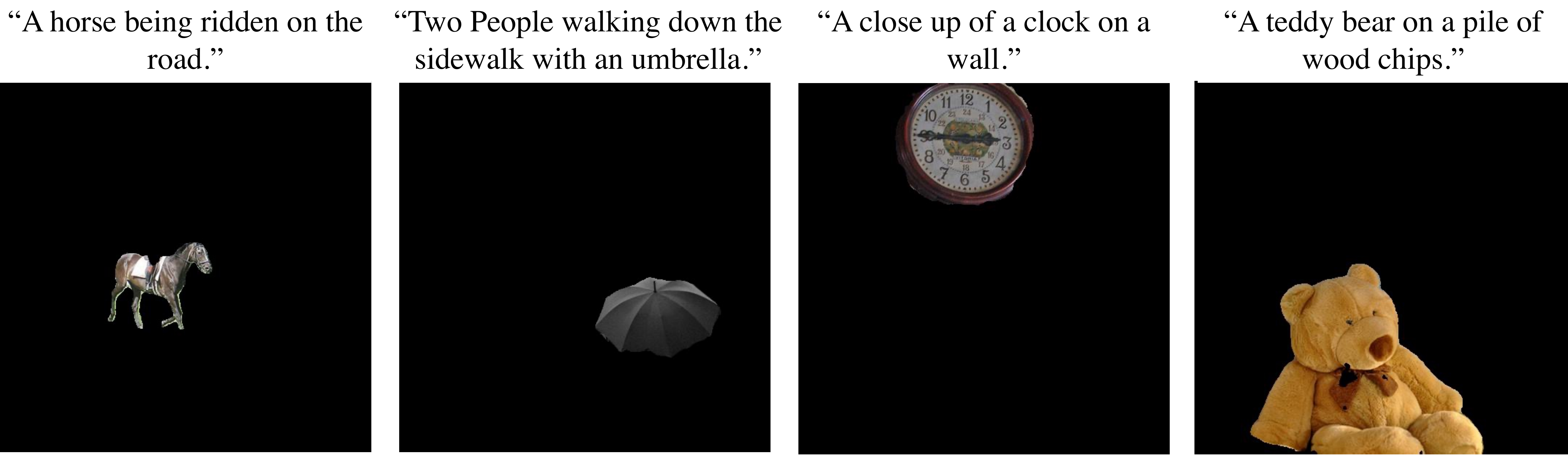} \\
    \end{tabular}
    
    \caption{Random examples from our subject-driven generation dataset, SubGen. Each example contains a caption and an image of a subject. The desired output is an image the aligns with both the subject and the caption.}
    \label{fig:subject-driven-comparison}
\end{figure*}

\section{Generated Examples By \name}
\begin{figure*}[t]
    \centering
    \setlength{\tabcolsep}{2.0pt}
    
        \begin{tabular}{c@{\hskip 1.3cm}c@{\hskip 1.0cm}c@{\hskip 1.4cm}c@{\hskip 2.3cm}c@{\hskip 1.5cm}c@{\hskip 1.5cm}c}
        & 
        & \makecell{\textbf{Conditioned Image}}
        & 
        & \makecell{\textbf{Generated Image}}
        & 
    \end{tabular}
    
    \begin{tabular}{cc}
    
        \includegraphics[width=0.67\textwidth]{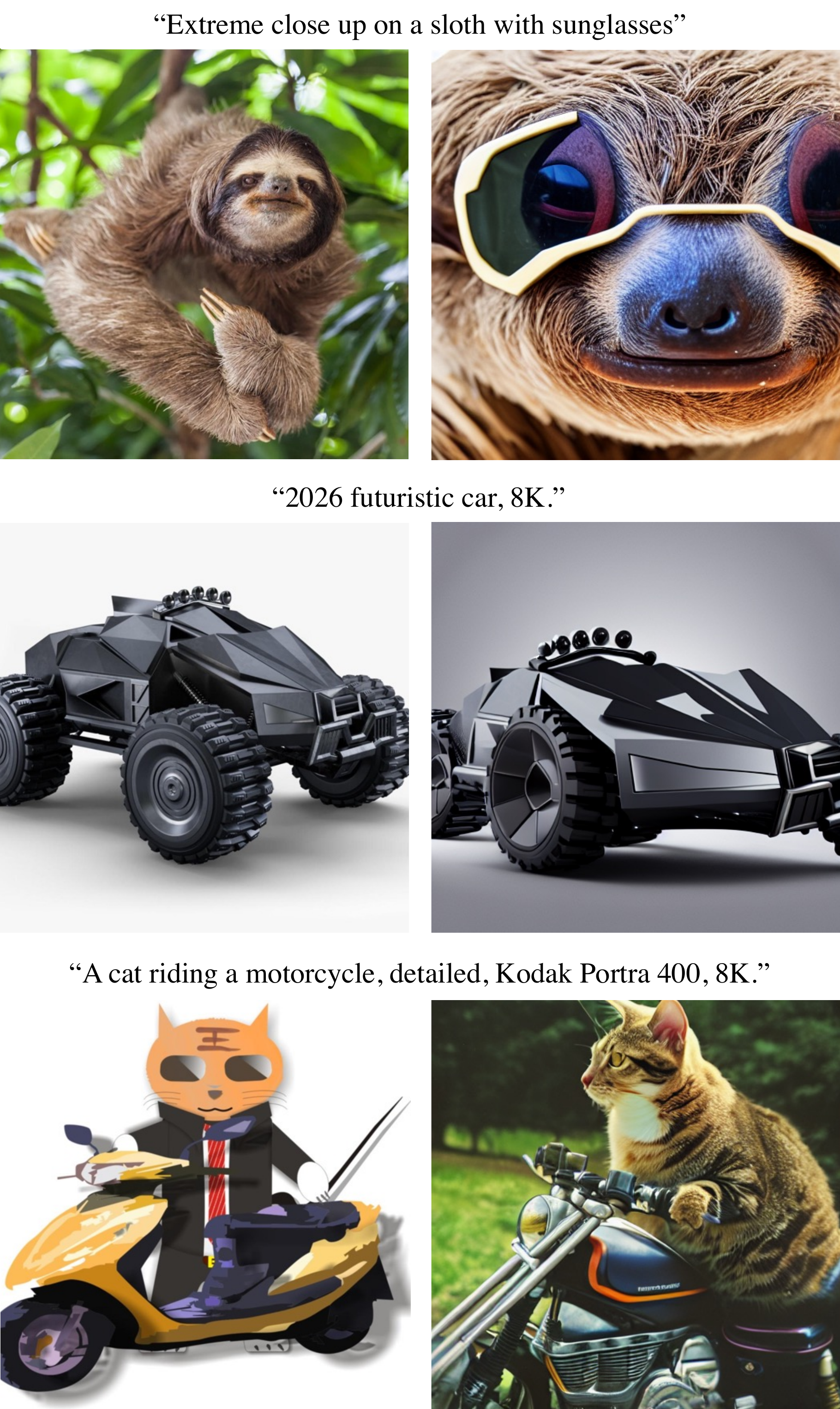} \\
        
    \end{tabular}
    
    \caption{Examples generated with Retrieve\&\method. We show the conditioned image (left) and the generated image (right).}
    \label{fig:retrieve-matrix}
\end{figure*}

\begin{figure*}[t]
    \centering
    \setlength{\tabcolsep}{2.0pt}
    
        \begin{tabular}{c@{\hskip 1.3cm}c@{\hskip 1.0cm}c@{\hskip 1.4cm}c@{\hskip 2.3cm}c@{\hskip 1.5cm}c@{\hskip 1.5cm}c}
        & 
        & \makecell{\textbf{Conditioned Image}}
        & 
        & \makecell{\textbf{Generated Image}}
        & 
    \end{tabular}
    
    \begin{tabular}{cc}
    
        \includegraphics[width=0.7\textwidth]{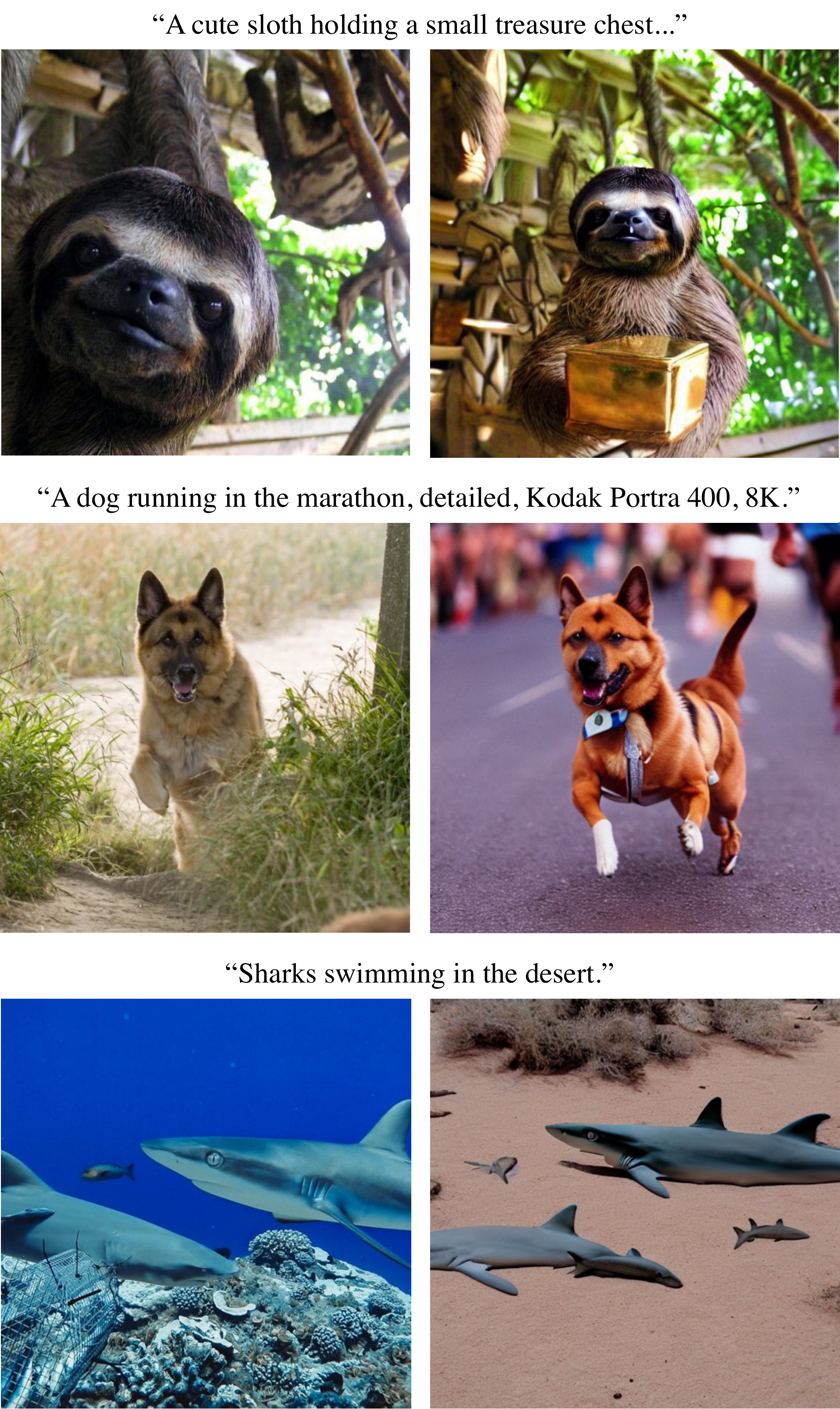} \\
        
    \end{tabular}
    
    \caption{Examples generated with Retrieve\&\method. We show the conditioned image (left) and the generated image (right).}
    \label{fig:retrieve-matrix-2}
\end{figure*}

\begin{figure*}[t]
    \centering
    \setlength{\tabcolsep}{2.0pt}
    \begin{tabular}{c@{\hskip 0.2cm}c@{\hskip 1.0cm}c@{\hskip 0.7cm}c@{\hskip 2.3cm}c@{\hskip 1.5cm}c@{\hskip 1.5cm}c}
        & \makecell{\textbf{Conditioned Image}}
        & 
        & 
        & \makecell{\textbf{Generated Images}}
        & 
    \end{tabular}
    \begin{tabular}{cc}
    
        \includegraphics[width=0.75\textwidth]{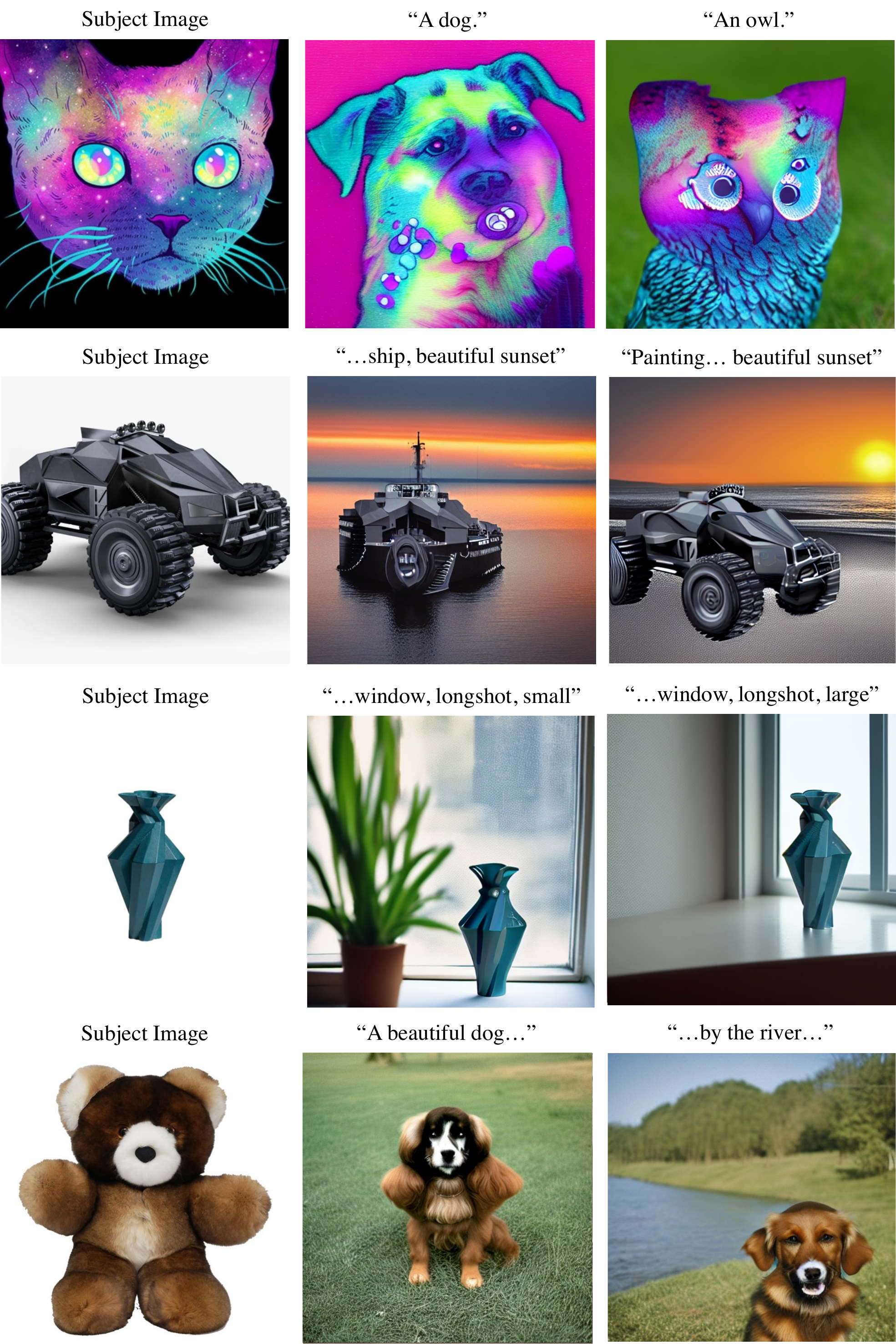} \\
        
    \end{tabular}
    
    \caption{Examples generated with Crop\&\method. We show the conditioned image (left) and generated images (middle and right).}
    \label{fig:crop-matrix}
\end{figure*}

\begin{figure*}[t]
    \centering
    \setlength{\tabcolsep}{2.0pt}
    \begin{tabular}{c@{\hskip 0.2cm}c@{\hskip 1.0cm}c@{\hskip 0.7cm}c@{\hskip 2.3cm}c@{\hskip 1.5cm}c@{\hskip 1.5cm}c}
        & \makecell{\textbf{Conditioned Image}}
        & 
        & 
        & \makecell{\textbf{Generated Images}}
        & 
    \end{tabular}
    \\
    \begin{tabular}{cc}
    
        \includegraphics[width=0.75\textwidth]{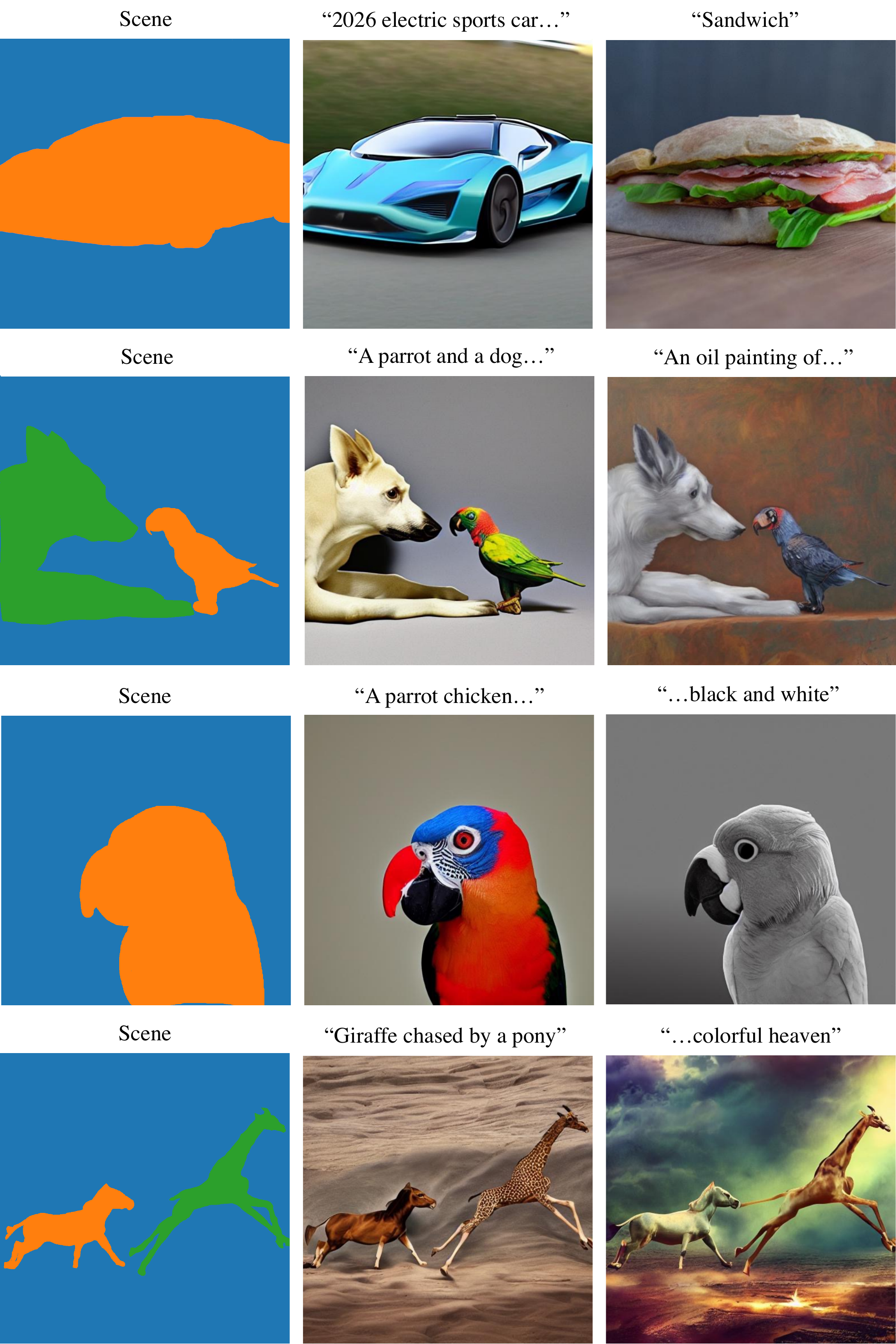} \\
        
    \end{tabular}
    
    \caption{Examples generated with Scene\&\method. We show the conditioned image (left) and generated images (middle and right).}
    \label{fig:scene-matrix}
\end{figure*}

\end{document}